\documentclass[balance,upint,subscriptcorrection,varvw,mathalfa=cal=boondoxo,spanish,french,vietnamese,russian,greek,pdf-a,colorlinks]{asmeconf}

\usepackage{multirow}
\usepackage{multicol}
\usepackage{gensymb}
\hypersetup{%
	pdfauthor={John H. Lienhard},									  
	pdftitle={ASME Conference Paper LaTeX Template},                  
	pdfkeywords={ASME conference paper, LaTeX template, BibTeX style},
	pdfsubject = {Describes the asmeconf LaTeX template},			  
	pdflicenseurl={https://ctan.org/pkg/asmeconf},
}

\begin{document}





\title{Multimodal Object Detection using Depth and Image Data for Manufacturing Parts} 
 
%
%
%






\SetAuthors{%
	Nazanin Mahjourian\affil{}\CorrespondingAuthor{mahjouri@mtu.edu}, 
	Vinh Nguyen\affil{}
}

\SetAffiliation{}{Department of Mechanical Engineering - Engineering Mechanics,\\
Michigan Technological University, Houghton, MI 49931}




\maketitle



\keywords{Computer Vision, Multimodal Object Detection, Early Fusion, Smart Manufacturing}


\begin{abstract}

Manufacturing requires reliable object detection methods for precise picking and handling of diverse types of manufacturing parts and components. Traditional object detection methods utilize either only 2D images from cameras or 3D data from lidars or similar 3D sensors. However, each of these sensors have weaknesses and limitations. Cameras do not have depth perception and 3D sensors typically do not carry color information. These weaknesses can undermine the reliability and robustness of industrial manufacturing systems. To address these challenges, this work proposes a multi-sensor system combining an red-green-blue (RGB) camera and a 3D point cloud sensor. The two sensors are calibrated for precise alignment of the multimodal data captured from the two hardware devices. A novel multimodal object detection method is developed to process both RGB and depth data. This object detector is based on the Faster R-CNN baseline that was originally designed to process only camera images. The results show that the multimodal model significantly outperforms the depth-only and RGB-only baselines on established object detection metrics. More specifically, the multimodal model improves mAP by 13\%  and raises Mean Precision by 11.8\% in comparison to the RGB-only baseline. Compared to the depth-only baseline, it improves mAP by 78\% and raises Mean Precision by 57\%. Hence, this method facilitates more reliable and robust object detection in service to smart manufacturing applications.

\end{abstract}






\section{Introduction}

Employing Artificial Intelligence (AI) for automation of manufacturing has resulted in increased efficiency, precision, and flexibility and created a paradigm shift in the design of manufacturing systems. AI has been successfully applied to a vast array of manufacturing tasks in the industry \cite{AI,AZIMIRAD2022319,ziad2024knowledge}. While AI-based methods have improved the manufacturing process, there are still challenges in ensuring that the AI-based black box systems continue to be reliable and robust. The most fundamental layer underlying all smart manufacturing systems is object detection, which allows the system to identify the type and position of the objects that it needs to handle. Object detection is a established computer vision problem, which involves identifying and categorizing specific objects of interest within a larger image by placing a bounding box around each detected object \cite{objectdetection}.

An effective automation system also requires proper sensor design to provide adequate coverage over the environment and allow the system to properly observe the scene and the objects \cite{maqsoodi2024framework, ziad2024pyramid}. The past decades have seen great advancement in sensor hardware. Camera resolutions have increased and they have become more affordable at the same time. Similarly, consumer applications have facilitated mass manufacturing of 3D sensors like lidars and stereo cameras, which are great sensors for smart manufacturing. Despite these improvements, sensors have inherent limitations rooted in their physics \cite{s23156997}. For example, a single image captured by camera does not carry depth information. 3D sensors capture point cloud data which addresses this issue, but these sensors are typically low resolution and do not provide color information.

An effective and reliable automation system requires selecting the right sensors and an object detection system that can effectively ingest the data provided by these sensors. Prior work has shown the limitations of object detection systems that rely solely on cameras~\cite{2Ddetection}. Image distortions like blur and noise can significantly lower the detection accuracy~\cite{blur}. Although cameras provide color information, there are environments where there is low contrast between objects and the background, and as a result detection accuracy may suffer~\cite{contrast} and cameras may not be enough for handling these environments. Moreover, camera-based object detection systems are sensitive to illumination and can be fragile if there are changes to the lighting conditions. It has also been shown that illumination can negatively affect~\cite{illumination} camera-based object detection systems, because they struggle to generalize to operate under lighting conditions different from what they have experienced during training. Similarly, the performance of these systems diminishes in scenarios where objects vary significantly in size or when they blend indistinguishable with the background in camera's view~\cite{background}. 

While 3D sensors are less sensitive to lighting, they pose their own set of challenges when used in industrial environments~\cite{photonics11030243}. As an example, object detection systems using 3D sensors often struggle when the scene contains densely arranged objects~\cite{3ddensley}. In addition, presence of objects with similar shapes and sizes, or objects with repetitive patterns cause difficulties for systems relying on point clouds for object detection~\cite{samesize}. Another challenge with using point cloud data is the extra complexity of modeling 3D information and the slower speed of processing 3D information. This is a significant obstacle, particularly for manufacturing environments which require real-time object detection and therefore cannot afford too much complexity and computational overhead~\cite{speed,Depth-estimation}. These conditions emphasize the need for advanced detection techniques that can take advantage of 3D sensors for improved performance while keeping the system simple and computationally efficient.

Since different sensors can be complementary and cover each other's blind spots and weaknesses, it makes sense to create object detectors which can leverage the strengths of multiple sensors to improve accuracy and dependability rather than relying on a single sensor. Multimodal object detection~\cite{multimodal, multimodal2}, which utilizes data from multiple sensor modalities has the potential to address the above-mentioned limitations of single-sensor systems. Integrating information from multiple sensors can be achieved via the sensor fusion process~\cite{sensorfusion}, which can produce more accurate and more reliable data for the object detection model. Different types of sensor fusion have been studied~\cite{typesoffusion} in the past.  Early fusion methods merge sensor data at the input stage. Intermediate fusion methods combine features derived from different sensors at an intermediate layer of a model. Lastly, late fusion approaches aggregate decisions proposed from different sensors at the final stage in the model. All types of sensor fusion can contribute to overcoming the limitations of single-sensor systems.

Prior work has explored multimodal object detection. One notable approach~\cite{feature} added a depth branch to the Faster-RCNN architecture to process depth in parallel with the RGB data. The depth branch created feature maps from depth inputs and these feature maps were concatenated with feature maps generated from RGB images. The authors showed that using depth allowed the model to succeed in some scenarios where color information alone could not distinguish objects from their backgrounds~\cite{feature}. Similarly, Zhu et al.~\cite{zhu2020} added a depth processing branch to the Faster R-CNN framework which enabled their model to handle both RGB and depth images for enhanced object detection. Their method employed two distinct CNNs to extract features from RGB and depth images independently. They also used depth information to delineate object edges more clearly and distinguish them from their backgrounds. Following feature extraction, the features were aligned and merged using a feature fusion layer, which implemented sum fusion, max fusion, and concatenation fusion strategies. Garbouge et al.~\cite{4channel} presented an approach that integrated RGB and depth data at an early stage by stacking them into a four-channel input for a CNN inspired by the AlexNet architecture for a classification task and showed that employing depth information improves object classification metrics.

However, none of the existing methods have explored efficient four-channel RGB+D inputs in the context of object detection tasks\cite{zhou2021rgb}. The processing of RGB and depth data through a single shared backbone offers several advantages. Firstly, when RGB and depth data are stacked at the input level, the model is able to efficiently extract features that merge information from both modalities, for example color and texture from RGB alongside spatial and structural details from depth. This integrated approach enables the network to learn more meaningful features early in the feature extraction process. By processing both data types simultaneously, the network can better generalize to unseen data as it learns to recognize and interpret patterns across both modalities. Additionally, employing a single backbone for processing reduces the computational overhead significantly. This computational efficiency comes from managing only one set of weights during back-propagation and inference. This approach creates a leaner model which is faster and less costly to run in a manufacturing plant. Lastly, the lower architectural complexity and lack of complex fusion mechanisms simplifies the development and tuning process for training and deploying the final model in the manufacturing environment.

This work introduces a novel method that employs early sensor fusion, where data from RGB images and 3D point clouds are integrated at the data level before being used by the detection algorithm. This method utilizes the combined strengths of both modalities for a more comprehensive representation of the scene that enhances object detection capabilities. Early fusion facilitates the extraction of comprehensive features that embody both the visual and spatial attributes of objects, which addresses the challenges posed by occlusions, variable sizes, and complex background information more effectively than late fusion or single-sensor methods. Hence, this work presents a novel model tailored for processing RGB-D images for object detection tasks in manufacturing. The rest of the paper is organized as follows. Section ~\ref{sec:method} discusses the methodology of the multimodal object detection framework. Section ~\ref{sec:experiments} outlines experiments which compare this approach to both RGB-only and Depth-only object detection of a modified NIST manufacturing task board. Section ~\ref{sec:results} discusses the experiment results, and Section ~\ref{sec:conclusions} discusses the conclusions and suggestions for future work.


\section{Multimodal Object Detection Model}
\label{sec:method}
\begin{figure*}[htb!]
    \centering
    \includegraphics[width=\linewidth]{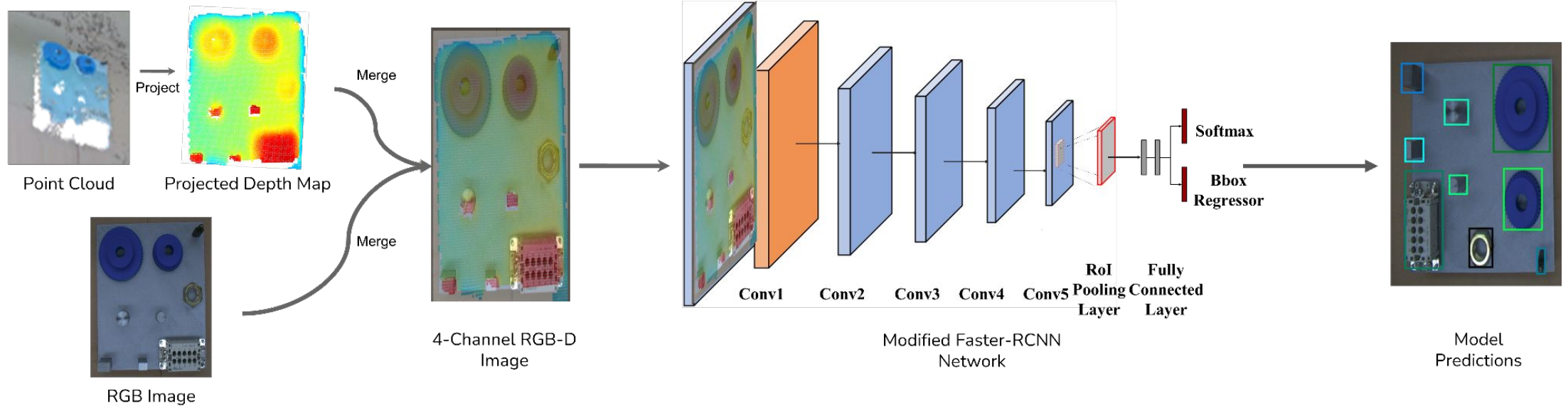}
    \caption{Overview of the RGBD-Man multimodal object detection framework.}
    \label{fig:overview}
\end{figure*}
Figure~\ref{fig:overview} illustrates the design of RGBD-Man, our multimodal object detection model for smart manufacturing. Our method combines data from two separate sensors which provide images and 3D point clouds. This two-sensor setup maximizes the generalizability of the method to be applicable to a wide range of manufacturing environments. 

The process starts with calibrating the two sensors so that the image and point cloud data are aligned when they are fed to the model. For each scene, the 3D point cloud is projected onto the image plane in 2D using the intrinsic and extrinsic projection matrices obtained through the sensor calibration process. This 3D to 2D projection converts the point cloud into a single-channel depth map. The depth map contains values which show each point's distance from the RGB camera. The depth map is then concatenated with the RGB image to form a four-channel input. This input is fed to a convolutional backbone which extracts feature maps from the combined RGB+depth data. Section ~\ref{sec:experiments} will outline the process for calibrating the camera and point cloud sensors and obtaining the depth maps using 3D-to-2D projections in more detail. This section discusses only the architecture of the object detector.

The object detection model is based on Faster R-CNN~\cite{fasterRCNN}, which is a popular deep learning architecture for its ability to efficiently localize and classify objects within images. A number of modifications are made to customize the Faster R-CNN model for accepting the fused RGB-D inputs. The first convolutional layer is replaced with a custom layer which accepts four input channels instead of three. The depth values are normalized and scaled to map to the range $0 - 255$ to match the scale of the RGB inputs. In addition, mean and standard deviation of depth values are computed over the dataset and used inside the model to normalize the depth inputs. Section ~\ref{sec:experiments} will outline more details about the training and evaluation setup, including the architecture of RGB-only and Depth-only variants of this object detection model.



\section{Experimental Setup}
\label{sec:experiments}

This section outlines the experimental setup for training and evaluating the multimodal object detection framework. First, the assembly task board and the manufacturing components are introduced. Then, the sensors and the sensor calibration process is discussed. Then, the steps for generating input depth maps are outlined. Next, three model variants are introduced to help us quantify the impact of depth inputs for object detection. Lastly, the setup and hyperparameters for training and evaluating these models are discussed.


\subsection{Assembly Task Board}

\begin{figure}[htbp!]
    \centering
    \includegraphics[width=0.7 \linewidth]{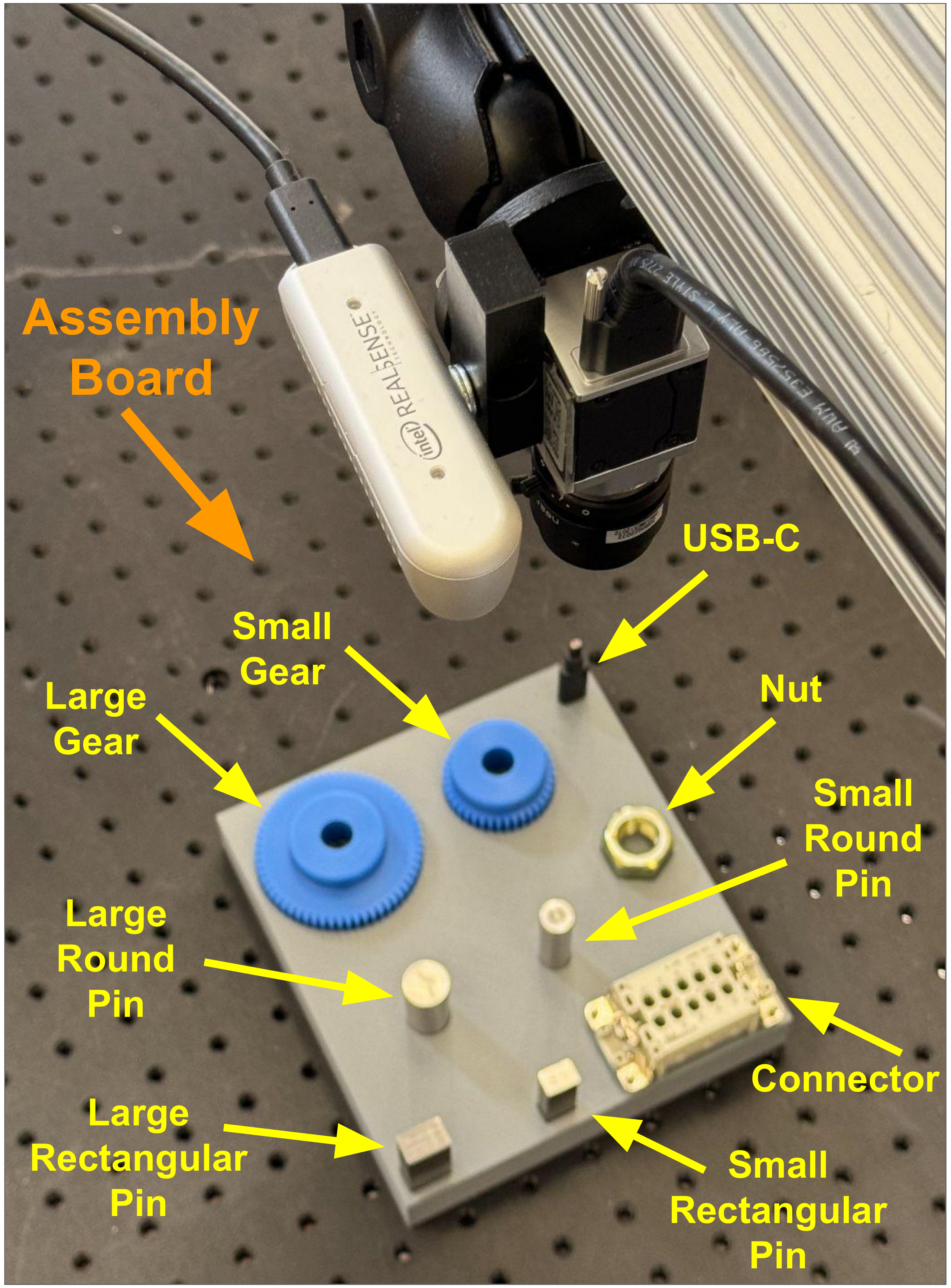}
    \caption{Assembly board and the manufacturing components.}
    \label{fig:board}
\end{figure}

The experiments were designed within the context of robotic pick-and-place applications, which is a common task in industrial manufacturing. The presented framework is relevant as it enables precise localization of the objects, which is critical for grasping and placement in challenging scenarios. The task board used in this work is derived from the modified NIST manufacturing task board configuration which contains nine components serving as the training and testing ground for the proposed object detection system. This board which is shown in Figure~\ref{fig:board} includes a large gear, a small gear, a USB-C connector, a nut, a waterproof connector, pairs of small and large rectangular pins, and pairs of small and large round pins. The small and large round pins have diameters of approximately 0.47" and 0.62", with corresponding exposed heights of 1.17" and 1.19", while the rectangular pins measure $0.39"\times 0.62"\times 1.19"$ for the large variant and $0.31"\times 0.47"\times 1.14"$  for the small variant. The gears also differ in size, with the large gear having a diameter of 2.42" and the small gear 1.64", with corresponding exposed heights of 0.69" and 0.66". The USB-C and waterproof connectors have irregular shapes, with bounding box dimensions of $0.25"\times 0.47"\times 1.45"$ for the USB-C connector and $1.27"\times 2.51"\times 0.99"$ for the waterproof connector. The hexagonal nut has a flat-to-flat width of 0.93", an inner diameter of 0.54", and exposed height of 0.43". The usage of this task board introduces realistic scenarios where lighting conditions, component orientation, and material finishes can affect the accuracy of object identification. While the experiments did not include varying lighting conditions, lighting effects were accounted for by incorporating shiny objects such as the four pins and the nut into the dataset. This selection of components provides a comprehensive framework for evaluating the capabilities of the system under study. Variations in surface texture, from gear smoothness to the metallic finish of connectors and pins, generate reflections and shadows that may obscure features or mimic other objects, risking misclassification. In addition, material properties including pin translucency and intricate details of components such as USB-C connectors which have very few points in space that represents its depth introduce visual variability that demands sophisticated interpretation by the network. These factors underscore the need for an advanced, adaptable neural network and thorough pre-processing to ensure accurate detection amidst the multifaceted visual conditions of the task board. 

\subsection{Sensors and Calibration}

\begin{figure}[htbp!]
    \centering
    \includegraphics[width=0.7 \linewidth]{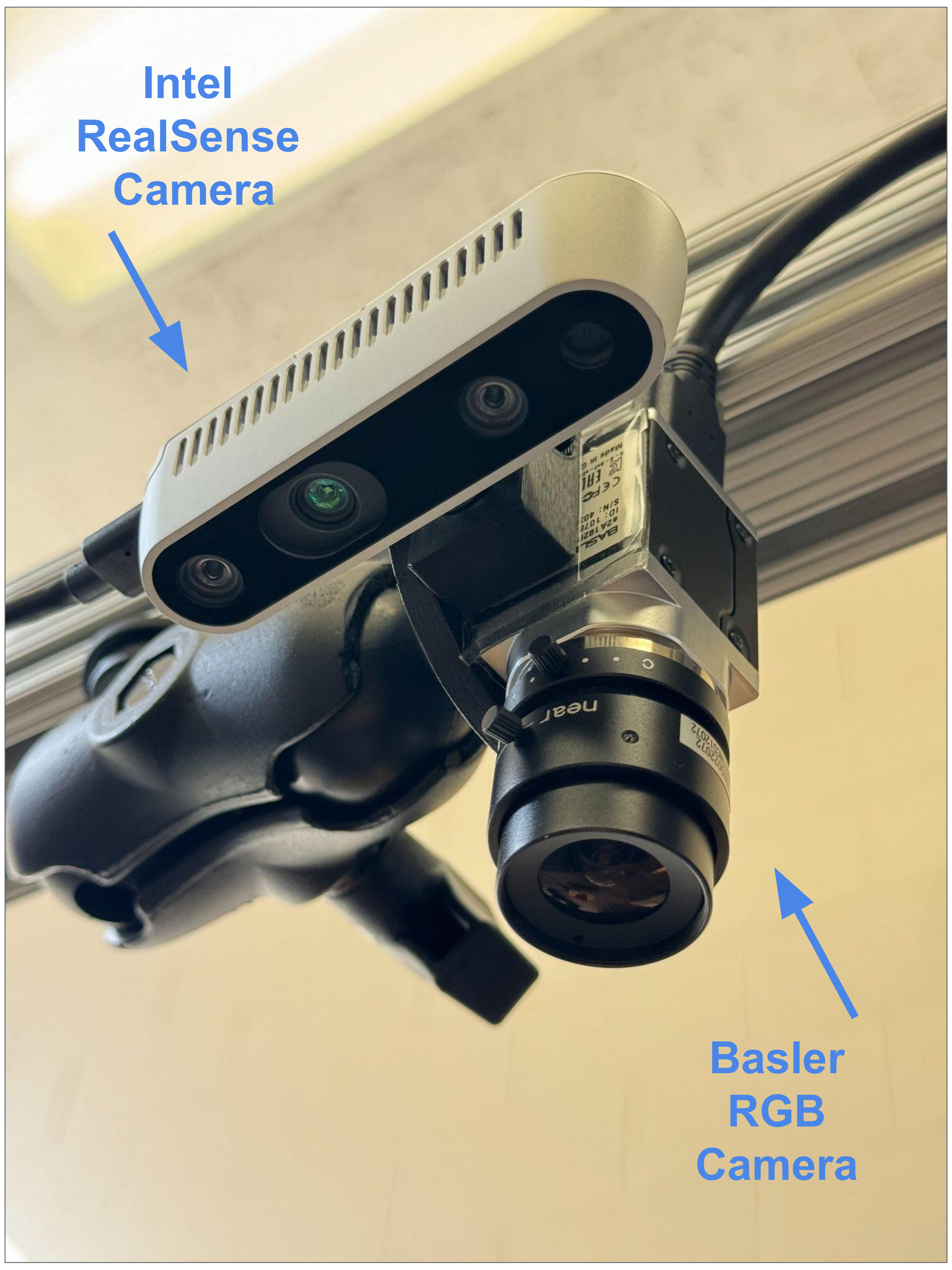}
    \caption{Multi-sensor setup featuring a Basler RGB camera and an Intel RealSense camera mounted to the robot's table. The location and orientation of the sensors remained fixed throughout all experiments.}
    \label{fig:sensor}
\end{figure}

The object detection framework ingests data from two independent sensors capturing image and depth data. Figure~\ref{fig:sensor} shows the multi-sensor setup used in our experiments.The sensors are mounted together and affixed to the robot's table. A Basler camera captures high resolution images to provide detailed texture, color, and shape information which are necessary for distinguishing objects based on color contrast against the background. An Intel RealSense depth camera senses the environment 3D in the form of point clouds to capture the spatial structure of the scene. This additional depth information allows the model to better differentiate objects and estimate their positions. This combination of sensors was selected to provide visual details of the scene in addition to the spatial relationships and dimensions of the objects under study. The separate Basler RGB camera was used to ensure higher image quality and flexibility in the multimodal setup. It has higher resolution and color accuracy in comparison to the RealSense RGB sensor. Although certain types of hardware such as RealSense used in this study can provide both RGB and depth data from a single sensor, our multi-sensor framework provides greater compatibility and generalizability to diverse manufacturing problems and environments. Using this approach any kind of sensor with any combinations can be integrated to satisfy needs of the manufacturing environment.

\begin{figure}[htbp!]
    \centering
    \includegraphics[width=\linewidth]{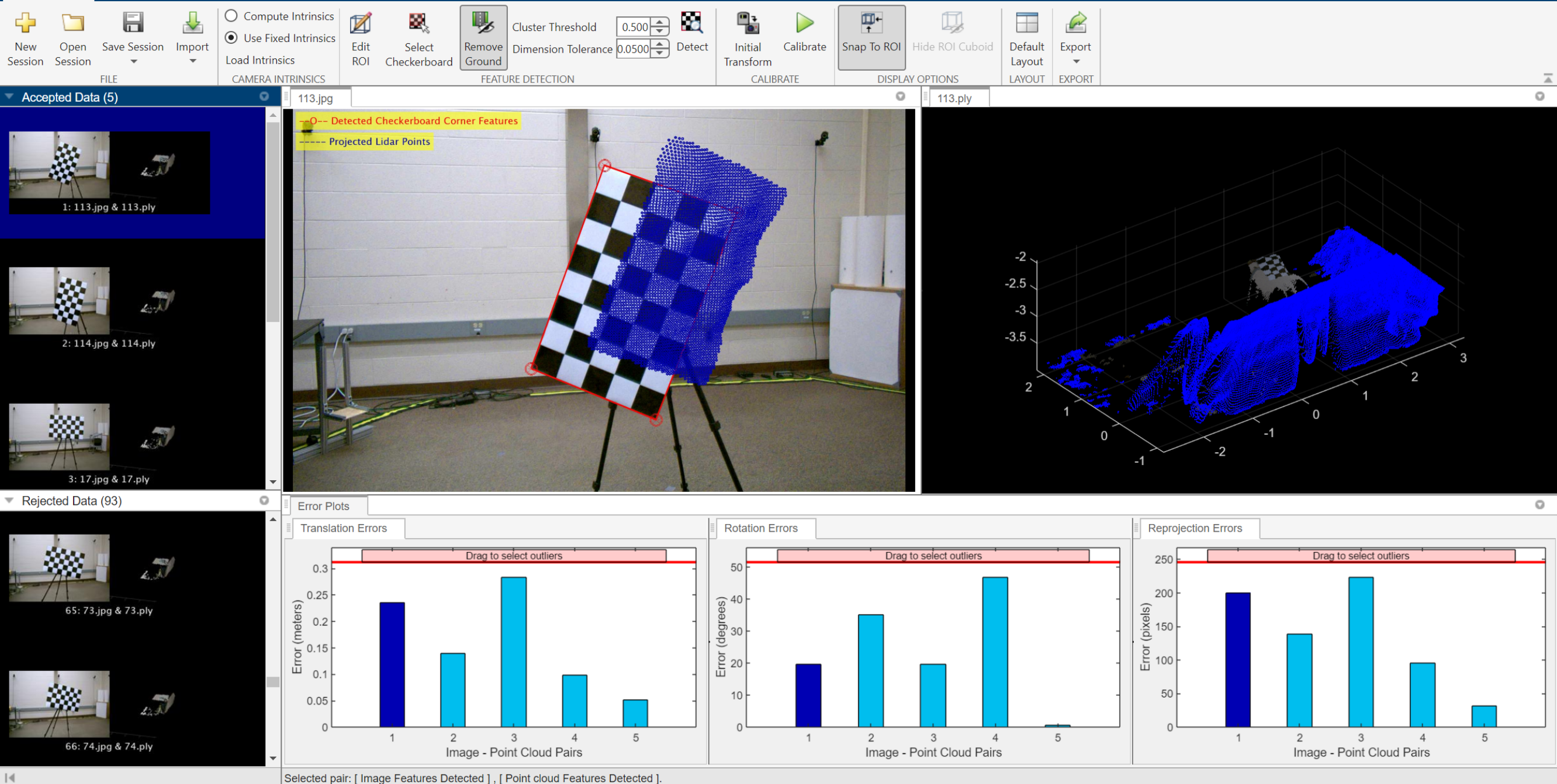}
    \caption{MATLAB's Lidar-Camera Calibration application output before 3D point cloud alignment.}
    \label{fig:calibration}
\end{figure}

The two sensors need to be calibrated against each other since the assembly task and the object detection model require precise alignment between the image and depth data received from the environment. Camera calibration~\cite{calibration} involves the precise estimation of camera parameters, including intrinsic and extrinsic, to infer accurate geometric features from captured sequences. To conduct the calibration, an asymmetric 10x7 checkerboard pattern was placed in various positions and orientations within the two sensor's field of view and 60 pairs of image and point clouds were captured. MATLAB's Lidar-Camera Calibration application~\cite{matlab_lidar_calibration} was used to tune the calibration parameters, as shown in Fig.~\ref{fig:calibration}. Fig.~\ref{fig:calibration} shows that the camera intrinsics are properly calibrated, therefore the red boundary perfectly lines up with the checkerboard's perimeter. However, the extrinsics are not calibrated yet, so the point cloud does not align with the checkerboard in the camera view. At the end of the process, the visual display and the error metrics signal a proper calibration. Hence, this process involved iteratively removing image and point cloud pairs with high calibration errors until the maximum translation error was below $0.0045m$ and the maximum rotation error was below $4.5\degree$. At this point, a total of 37 image - point cloud pairs were used to compute the calibration parameters.


\subsection{Generating Depth Maps}


The 3D sensor captures point clouds, which are sets of disjoint points in 3D space. For higher efficiency, our multimodal object detector consumes depth maps, which are generated from the point clouds. A depth map is essentially a single-channel image similar to a grayscale image. Unlike a grayscale image where each pixel indicates a shade of gray, each pixel in a depth map specifies the distance from the sensor to the object existing at that pixel. If a pixel has no corresponding points, it will record a value of zero.

The point cloud data is projected onto the 2D image plane using a $4 \times 4$ homogeneous transformation matrix obtained from sensor calibration. This transformation matrix produces metric depth values. To make the values easier to consume in the model, they are normalized as

\begin{equation} \label{eq_1}
    D_{norm} = \frac{D - D_{min}}{D_{max} - D_{min}}
\end{equation}
\\
where $D$ is the original depth value, $D_{min}$ and $D_{max}$ are the minimum and maximum depth values observed across the entire dataset, and $D_{norm}$ is the normalized depth value in range $[0, 1]$.

Since the model receives RGB images encoded as unsigned integers in range $[0, 255]$, this work further scales the depth values as

\begin{equation} \label{eq_2}
    D_{s} = D_{norm} \times 255
\end{equation}
\\
where $D_s$ is the scaled depth value. After this transformation, the depth map is concatenated with the raw RGB image to form a four-channel image where all values range consistently from 0 to 255, which makes the data suitable for feeding to the model. The four-channel RGB-D images can be conveniently precomputed and stored to disk since many image formats like PNG support a fourth alpha channel mainly used for encoding transparency information for each pixel.

The mean and standard deviation of depth values across the dataset were computed and used to renormalize the depth channel inside the object detection model similarly to the normalization of RGB channels. This renormalization increases the training efficiency of the model since it changes the inputs to small positive and negative values around zero.


\subsection{Data Collection and Labeling}

A dataset containing 301 pairs of images each in 1920 pixels in width and 1200 pixels in height, along with associated point clouds captured from varied configurations of the assembly components on and off the assembly board were created. Placing the objects on and off the board enables the robot to detect and localize the components both before and after they get installed on the assembly board.  The position and orientation of the assembly board within the sensors' field of view were also varied so that the model can generalize to varied task configurations. It was also ensured that each class of components is adequately present in the dataset to avoid an imbalance between classes~\cite{classimbalance} leading to suboptimal performance.

The Roboflow annotation tool~\cite{roboflow2022} was used to label every one of the 301 examples in the dataset, which were then stored into the COCO JSON format. The examples in the dataset were randomly split between a train set with 226 examples, a validation set with 45 examples, and a test set with 30 examples. After annotating the full RGB-D dataset, special variants of the dataset containing only image data (RGB-only) and only depth data (Depth-only) were made to enable training model variants for ablations discussed in the next section. All dataset variants shared the original class and bounding box labels and used the same train/validation/test split.


\subsection{Model Variants}

\begin{figure*}[htb!]
    \centering
    \includegraphics[width=\linewidth]{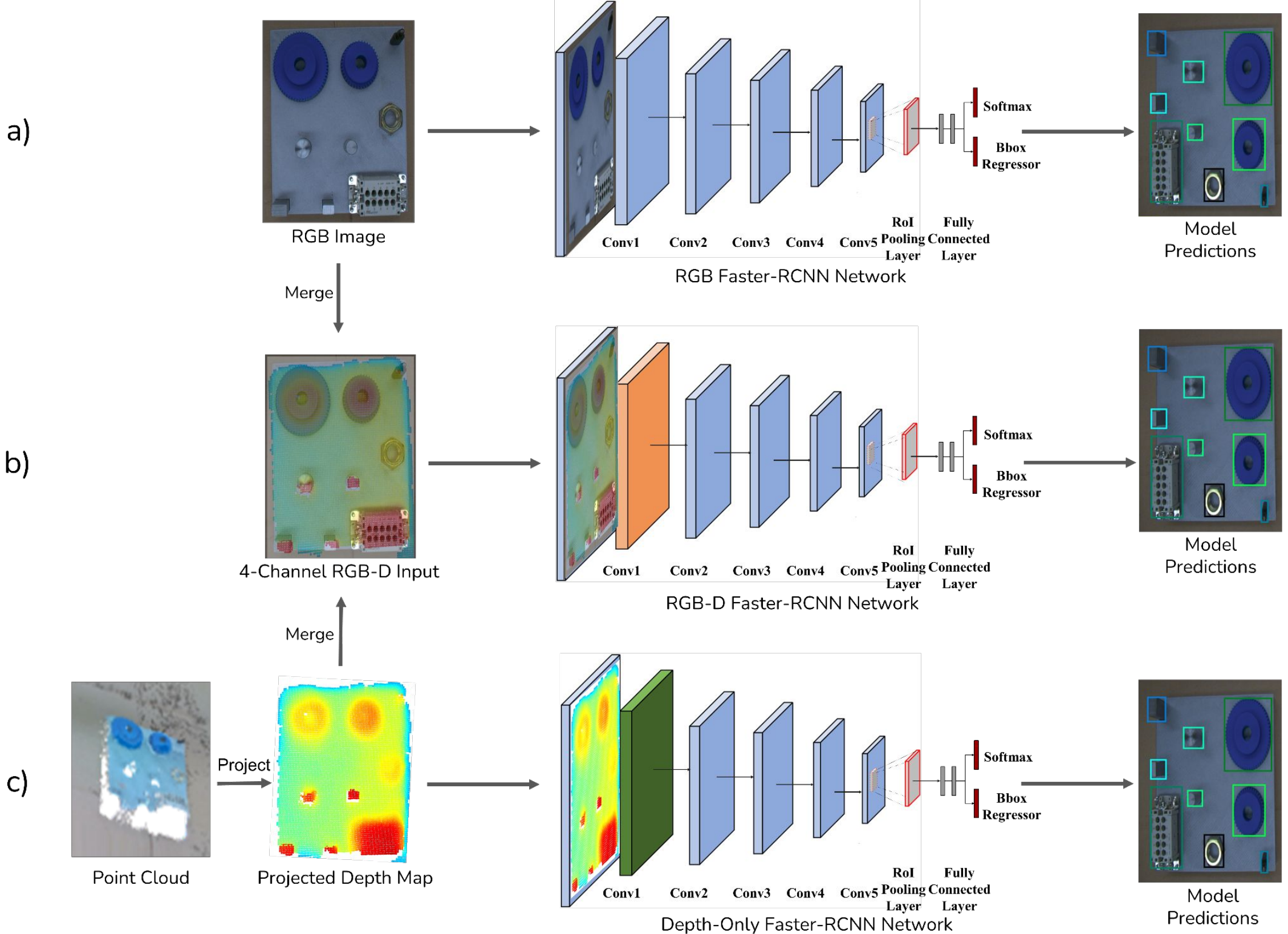}
    \caption{The three variants of the object detection model used in experiments. \emph{a.} RGB-only: The model receives only camera images and predicts object class scores and bounding boxes for the detected objects. \emph{b.} RGBD-Man: The main model which concatenates RGB and depth information into a four-channel input and predicts objects from the fused inputs. \emph{c.} Depth-only: The model which predicts objects given only a single-channel depth map of scene.}
    \label{fig:variants}
\end{figure*}

To help isolate the impact of feeding depth data to the object detection model alongside the images, two additional variants of object detection model were created. The first variant, called RGB-only, receives only camera data and predicts object class scores and bounding boxes. The second variant, called Depth-only, receives only single-channel depth maps and predicts the same outputs. These two variants are unimodal since they each receive a single input modality. These two variants, alongside the original multimodal RGBD-Man model form three distinct models that are fully trained and evaluated in our experiments. Fig.~\ref{fig:variants} compares the three variants. Each model variant is trained and evaluated on the corresponding dataset variant. Other than the first convolutional layers, all three variants shared the same architecture, built on top of a ResNet-50~\cite{resnet50} backbone. 

In all experiments, models are trained from scratch with all weights initialized to random values. No weights are shared between the three model variants, and no weights are initialized from pretraining on any other dataset.

\subsection{Training and Evaluation Setup}

All models were trained on NVIDIA Tesla T4 GPUs with a driver version of 535.104.05 and CUDA version 12.2. All models used a batch size of 4, which was a compromise between ensuring sufficient gradient stability and keeping the memory requirements within the limits of our hardware. The models were trained using the Stochastic Gradient Descent (SGD) optimizer with a learning rate of 0.001, Nesterov momentum of 0.9, and weight decay of 1e-4. SGD has been widely used in object detection tasks and has proven effective in training Faster R-CNN models. The SGD optimizer with the specified learning rate and Nesterov momentum helps the model converge to an appropriate local minimum while also benefiting from faster updates due to the momentum term. The SGD optimizer is computationally efficient, and can lead to better generalization performance compared to other optimizers like Adam.

The models were evaluated using standard objection detection metrics, namely mean Average Precision (mAP) at 0.5 IoU threshold, and Mean Precision~\cite{mAP}. The 0.5 IoU threshold for mAP is a widely accepted metric for evaluating the model performance in object detection benchmarks. This threshold provides a balance between localization accuracy and detection tolerance. It ensures that detected bounding boxes reasonably overlap with ground truth annotations.

An early stopping mechanism was used to allow each model variant to train for as long as it can improve the validation metrics. At the end of each training epoch, the latest model checkpoint were evaluated against the validation set using the mean Average Precision (mAP) metric. Whenever an improvement in mAP was detected, which corresponds to better model accuracy, the model's checkpoint was saved as the new best configuration. The training process was stopped if the metrics did not improve over a consecutive span of 10 epochs. However, the checkpoint with the best evaluation metrics was always saved and used for a final evaluation over the test set, which produced the final mAP and Mean Precision metrics for that variant. This adaptive strategy helps avoid overfitting to the training set, and validates the chosen hyperparameters, including batch size and optimizer. The consistency of the results across multiple runs further supports the reliability of the training setup.

To ensure reproducibility and robustness of the analysis, all three variants were trained and tested 10 times. This approach was chosen due to the random initialization of the model weights that could lead to variability in performance. By conducting multiple runs, the consistency of the results can be confirmed to validate the robustness of the proposed method. The repeated training sessions confirmed that the enhancements observed with the integration of the depth channel were indeed reproducible and not a result of favorable random weight initialization.

\begin{figure}[htbp!]
    \centering
    \includegraphics[width=\linewidth]{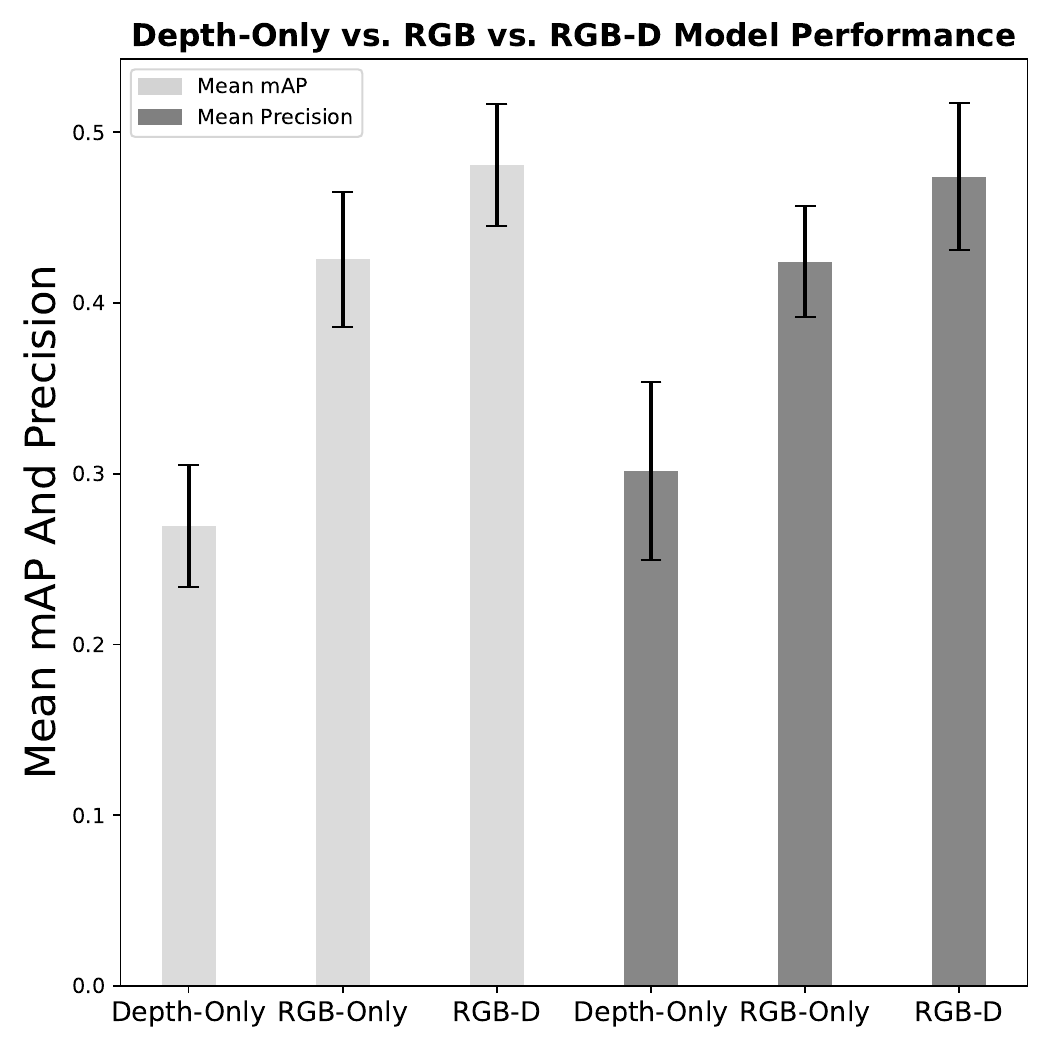}
    \caption{Comparison of the Depth-only, RGB-only, and RGB-D model variants over mAP and Mean Precision metrics on the test set. The error bars show standard deviation of results over 10 runs.}
    \label{fig:results_chart}
\end{figure}
\section{Results}
\label{sec:results}

\begin{figure*}[htb!]
    \centering
    \includegraphics[width=\linewidth]{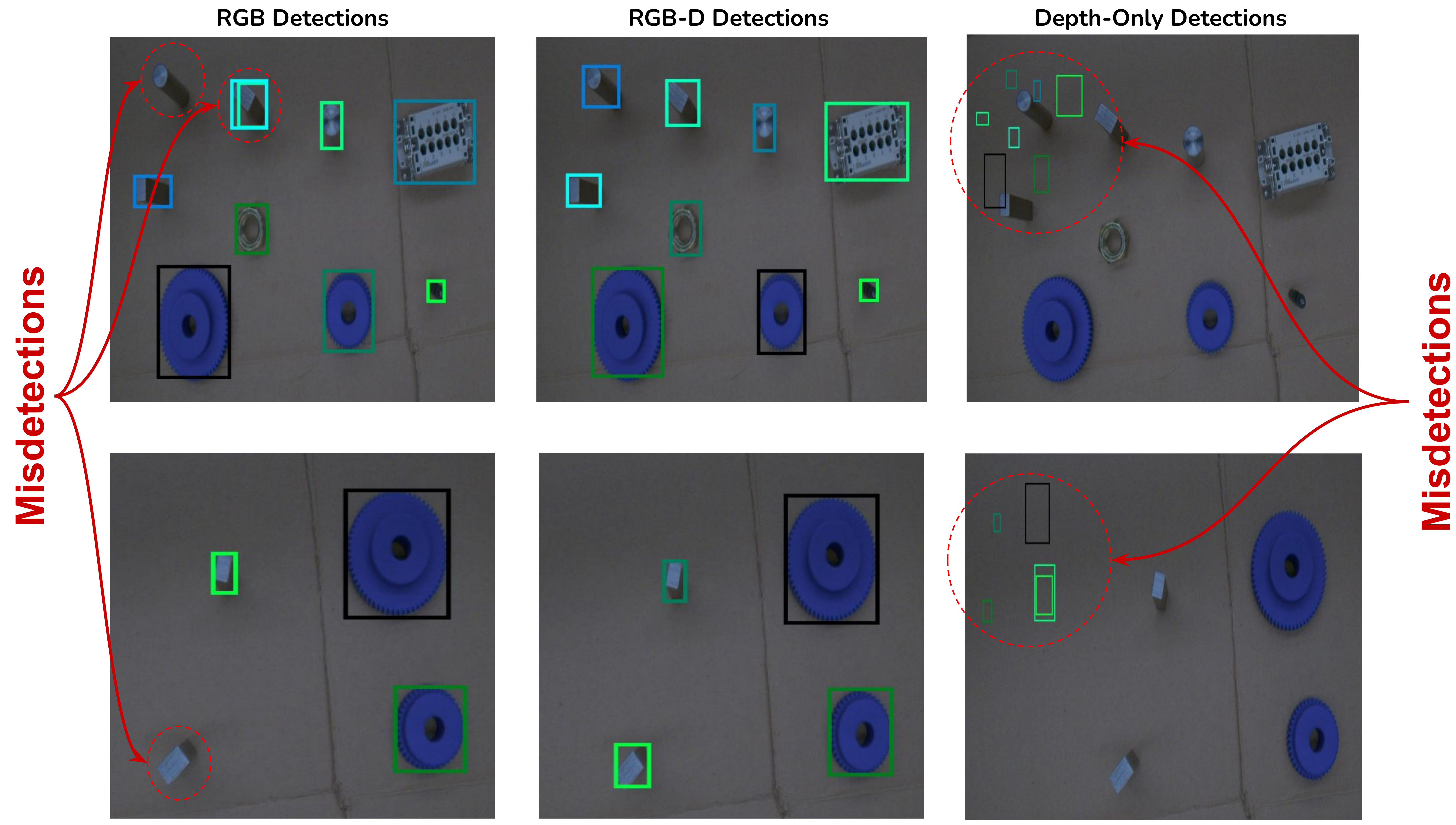}
    \caption{Qualitative comparison of the three model variants over two sample scenes from the test set. The RGB-only model tends to fail to detect metallic objects whose color hues are close to the background. The Depth-only model tends to fail detecting thin objects which may not have enough points captured in the 3D point cloud, and it may hallucinate detecting non-existing objects. The RGB-D model shows superior detection accuracy in these examples.}
    \label{fig:results_examples}
\end{figure*}
This section presents the evaluations results comparing the performance of the RGB-only, Depth-only, and RGB-D model variants using the mAP and Mean Precision metrics over the test set. Table~\ref{tab:results} and Fig.~\ref{fig:results_chart} compare the metrics from the three models. Fig.~\ref{fig:results_chart} also shows error bars indicating standard deviation of the metrics over the 10 runs for each model.

\begin{table}[htbp]
  \centering
  \begin{tabular}{|c|c|c|}
    \hline
    \textbf{Model} & \textbf{Mean mAP} & \textbf{Mean Precision} \\ \hline
    Depth-only & 0.269 & 0.301 \\ \hline
    RGB-only & 0.425 & 0.424 \\ \hline
    RGB-D & \textbf{0.480} & \textbf{0.474} \\ \hline
    
  \end{tabular}
  \caption{Comparison of the Depth-only, RGB, and RGB-D model variants over the test set.}
  \label{tab:results}
\end{table}

The RGB-D model, which uses both images and depth, achieves a mean mAP of 0.480, outperforming the RGB-only model's mean mAP of 0.425, and the Depth-only model's mean mAP of 0.269.  Similarly, the mean precision metric further demonstrates the RGB-D model's superior performance with a score of 0.474 in contrast to the RGB-only model's 0.424 and the Depth-only model's 0.301.  These results are visually summarized in Figure~\ref{fig:results_chart}, where the RGB-D model demonstrates a clear advantage over the RGB-only and depth-only models across both metrics. Specifically, the RGB-D model's mean mAP is 13\% higher than that of the RGB-only model and 78\% higher than the depth-only model. Similarly, the RGB-D model's mean precision shows an increase of 11.8\% compared to the RGB-only and a significant 57\% enhancement when measured against the Depth-only model. This demonstrates a pronounced enhancement in object detection capabilities when depth information is employed in conjunction with RGB data.

Figure~\ref{fig:results_examples} shows the performance of three model variants on two example scenes from the test set arranged side by side for a direct comparison. The qualitative results reinforce the quantitative findings. The RGB-D detections show a notable improvement in identifying objects that the RGB-only model misses--typically metallic objects whose colors are not very different from the scene background. The RGB-only model also struggles with reflective surface objects and objects with low contrast with the background. Metal pins often are missed in detection because reflection causes inconsistent pixel values. These corresponding RGB values change with different orientations and lighting, which lead to misdetections, as the model cannot reliably differentiate object boundaries. Objects that are misdetected or completely undetected in the RGB-only model are accurately identified by the RGB-D model. This visual comparison aligns with the quantitative results and emphasizes how the integration of depth information with RGB data enhances detection accuracy. The depth component aids in overcoming the limitations observed with RGB-only detections, particularly in complex scenarios where depth cues are crucial for accurate object localization and classification. The Depth-only model's performance is notably weaker compared to the RGB-only and RGB-D models. The Depth-only model tends to fail detecting objects which are short in height and small objects which may not be captured by many points in the point cloud from the 3D sensor. Note that the model lacks a full 3D representation or point cloud data. Only the depth values for every pixel in each image was given to the model in the shape of a one dimensional array. This results in the model lacking texture, shape, and edge information that are critical for object recognition. While depth information can be crucial for adding spatial context in a multimodal setting, it is not sufficient to reliably identify and localize objects, particularly in complex scenes or where depth data lacks the necessary detail and contrast to differentiate objects from the background.

The enhanced performance of the RGB-D model can be attributed to the additional spatial cues provided by the depth data, which complement the visual cues from the RGB data.  This extra information allows the RGB-D model to better interpret the three-dimensional structure of the scene, leading to improved detection of objects that may be challenging to recognize based solely on their appearance in RGB images.


\section{Conclusions}
\label{sec:conclusions}

This work demonstrates a notable advancement in the field of object detection for manufacturing by presenting an enhanced multimodal approach that combines RGB and depth data in an efficient manner. This work developed and validated an object detection model using an early fusion of depth information with standard RGB data that results in an efficient four-channel input that can be processed using standard convolutional networks. The experiments have demonstrated that the integration of depth information with RGB data can significantly improve object detection performance and advance the capabilities of object detection systems, particularly in complex manufacturing environments where visual ambiguity and presence of varied objects and components can pose problems for image-only object detection systems.

Although there is still a gap between the current results and the high accuracy required for industrial application, the performance would likely improve by utilizing a larger dataset with more diverse object configurations, lighting conditions, and occlusion scenarios. Additionally, a better sensor calibration and refined depth processing alongside with transfer learning and fine tuning the model on industrial settings will also help improve the accuracy to meet real world manufacturing requirements. Future work will explore scenarios involving transparent, irregular, and non-standard shapes, as well as fully 3D and overlapping objects to further investigate the role of depth in enhancing object detection in manufacturing environments. Moreover, future work will integrate and evaluate this object detection framework in a robotic object manipulation task~\cite{s24082585}. As future work, controlled robotic grasping experiments will be performed to determine if the current detection performance is sufficient for this robotic grasping task, or whether further refinements are necessary. Alternative designs for multimodal object detection will also be explored and evaluated, including the augmentation of 3D detection models with RGB data from cameras.




\nocite{*}

\bibliographystyle{asmeconf}  
\bibliography{asmeconf-sample}

\end{document}